\documentclass[11pt,a4paper]{article}
\usepackage[hyperref]{acl2020}
\usepackage{times}
\usepackage{latexsym}

\usepackage[T1]{fontenc}

\usepackage{amsmath}
\usepackage{amsfonts}
\usepackage{hyperref}

\usepackage{natbib}

\usepackage{booktabs, multirow} % for borders and merged ranges
\usepackage{soul}% for underlines
\usepackage{changepage,threeparttable} % for wide tables
\usepackage{graphicx}
\usepackage{subcaption}

\usepackage{microtype}

\usepackage[utf8]{inputenc}
\usepackage{microtype}

\aclfinalcopy % Uncomment this line for the final submission
\title{Comparing in context: Improving cosine similarity measures with a metric tensor}

\author{Isa M. Apallius de Vos \Thanks{These authors contributed equally to this work.} \\ \texttt{\{i.m.apalliusdevos\}}  \\ {\bf Ghislaine L. van den Boogerd \footnotemark[1]}\\ \texttt{\{g.l.vandenboogerd\}}  \\ {\bf Mara D. Fennema \footnotemark[1]} \\ \texttt{\{m.d.fennema\}} \\ 
\texttt{@students.uu.nl} \\
Utrecht University, The Netherlands \\\And Adriana D. Correia \Thanks{Corresponding author.}  \\ \texttt{a.duartecorreia@uu.nl} \\ Utrecht University, The Netherlands}

\date{}

\begin{document}
\maketitle
\begin{abstract}
%[review language model verbiage]
%[s_human]
%[]

Cosine similarity is a widely used measure of the relatedness of pre-trained word embeddings, trained on a language modeling goal. Datasets such as WordSim-353 and SimLex-999 rate how similar words are according to human annotators, and as such are often used to evaluate the performance of language models. Thus, any improvement on the word similarity task requires an improved word representation. In this paper, we propose instead the use of an extended cosine similarity measure to improve performance on that task, with gains in interpretability. We explore the hypothesis that this approach is particularly useful if the word-similarity pairs share the same context, for which distinct contextualized similarity measures can be learned. 
We first use the dataset of Richie et al. (2020) to learn contextualized metrics and compare the results with the baseline values obtained using the standard cosine similarity measure, which consistently shows improvement. We also train a contextualized similarity measure for both SimLex-999 and WordSim-353, comparing the results with the corresponding baselines, and using these datasets as independent test sets for the all-context similarity measure learned on the contextualized dataset, obtaining positive results for a number of tests.
\end{abstract}

\section{Introduction}

Cosine similarity has been largely used as a measure of word relatedness, since vector space models for text representation appeared to automatically optimize the task of information retrieval \cite{salton1983introduction}. While other distance measures are also commonly used, such as Euclidean distance \cite{witten2005practical}, for cosine similarity only the vector directions are relevant, and not their norms. More recently, pre-trained word representations, also referred to as embeddings, obtained from neural network language models, starting from word2vec (W2V) \cite{mikolov2013distributed}, emerged as the main source of word embeddings, and are subsequently used in model performance evaluation on tasks such as word similarity \cite{toshevska2020comparative}. Datasets such as SimLex-999 \cite{hill2015simlex} and WordSim-353 \cite{finkelstein2001placing}, which score similarity between word-pairs according to the assessment of several humans annotators, have become the benchmarks for the performance of a certain type of embedding on the task of word similarity \cite{recski2016measuring,dobo2020comprehensive, speer2017conceptnet, banjade2015lemon}.

For $\vec{n}_a$ and $\vec{n}_b$, the vector representations of two distinct words $w_a$ and $w_b$, cosine similarity takes the form

\begin{equation}\label{cosine}
cos_{ab}= \frac{\vec{n}_a\cdot \vec{n}_b}{|| \vec{n}_a|| \: ||\vec{n}_b||},
\end{equation} with the Euclidean \textit{inner product} between any two vectors $\vec{n}_a$ and $\vec{n}_b$ given as

\begin{equation} \label{euclideaninner}
\vec{n}_a \cdot \vec{n}_b = \sum_{i} \vec{n}^i_a \vec{n}^i_b,
\end{equation} and the \textit{norm} of a vector $\vec{n}_a$ given as

\begin{equation} 
    ||\vec{n}_a||= \sqrt{\vec{n}_a\cdot\vec{n}_a},
\end{equation} dependent on the inner product  \cite{axler1997linear}.

Using this measure of similarity, improvements can only take place if the vectors that represent the words change. However, the assumption that the vectors interact using a Euclidean inner product becomes less plausible when it comes to higher order vectors. If, differently, we consider that the vector components are not described in a Euclidean basis, then we enlarge the possible relationships between the vectors. Specifically in the calculation of the inner product, on which the cosine similarity depends, we can use an intermediary \textit{metric} tensor. By challenging the assumption that the underlying metric is Euclidean, cosine similarity values can be improved \textit{without changing vector representations}.

We identify two main motivations to search for improved cosine similarity measures. The first motivation has to do with the cost of training larger and more refined language models \cite{bender2021dangers}. By increasing the performance on a task simply by changing the evaluation measure without changing the pre-trained embeddings, we expect that better results can be achieved with more efficient and interpretable methods. This is particularly true of contextualized datasets, with benefits not only for tasks such as word similarity, but also others that use cosine similarity as a measure of relatedness, such as content based recommendation systems \cite{schwarz2017analysis}, and where it can be particularly interesting to explore the different metrics that emerge as representations of vector relatedness. 

The second motivation comes from compositional distributional semantics, where words of different syntactic types are represented by tensors of different ranks, and representations of larger fragments of text are produced via tensor contraction \cite{coecke2010mathematical,grefenstette2011experimental,grefenstette2011experimenting,milajevs2014evaluating,baroni2014frege,paperno2014practical}. This framework has proved to be a valuable tool for low resource languages, enhancing the scarce available data with a grammatical structure for composition, providing embeddings of complex expressions \cite{abbaszadeh2021parametrized}. As these contractions depend on an underlying metric that is usually taken to be Euclidean, improvements have only been achieved, once again, by modifying word representations \cite{wijnholds2019evaluating}. As proposed by \citet{correia2020density}, another way to improve on these results consists in using a different metric to mediate tensor contractions. Metrics obtained in tasks such as word similarity can be transferred to tensor contraction, and thus we expect this work to open new research avenues on the compositional distributional framework, providing a better integration with (contextual) language models.

This paper is organized as follows. In $\S$\ref{model} we introduce an extended cosine similarity measure, motivating the introduction of a metric on the hypothesis that it can optimize the relationships between the vectors. In $\S$\ref{methods} we explain our experiment on contextualized and non-contextualized datasets to test whether improvements can be achieved. In $\S$\ref{results} we present the results obtained in our experiments and in $\S$\ref{conclusion} we discuss these results and propose further work.

Our contributions are summarized below:
\begin{itemize}

\item Use of contextualized datasets to explore contextualized dynamic embeddings and evaluate the viability of contextualized similarity measures;

\item Expansion of the notion of cosine similarity, motivating our model theoretically, contributing to a conceptual simplification that yields interpretable improvements.

\end{itemize}

\subsection{Related Literature}

Variations on similarity metrics on the contextualized dataset of \citet{richie2020spatial} have been first explored in \citet{richie2021similarity}, but only on static vector representations and diagonal metrics. Other analytical approaches to similarity learning have been identified in \citet{kulis2013metric}. The notion of soft cosine similarity of \citet{sidorov2014soft} presents a relevant extension theoretically similar to ours, but motivated and implemented differently. Using count-base vector space models with words and n-grams as features, the authors extract a similarity score between features, using external semantic information, that they use as a distance matrix that can be seen as a metric; however, they do not implement it as in Eq. (\ref{generalinner}), but instead they transform the components by creating a higher dimensional vector space where each entry is the average of the components in two features, multiplied by the metric, whereas we, by contrast, learn the metric automatically and apply it to the vectors directly. \citet{hewitt2019structural} also use a modified metric for inner product to probe the syntactic structure of the representations, showing that syntax trees are embedded implicitly in deep models’ vector geometry.

Context dependency in how humans evaluate similarity, which we based our study on, has been widely supported in the psycholinguistic literature. \citet{tversky1977features} shows that similarity can be expressed as a linear combination of properties of objects, \citet{barsalou1982context} looks at how context-dependent and context-independent properties influence similarity perception, \citet{medin1993respects} explore how similarity judgments are constrained by the very fact of being requested, and \citet{goldstone1997similarity} test how similarity judgments are influenced by context that can either be explicit or perceived.

\section{Model}\label{model}

A metric is a tensor that maps any two vectors to an element of the underlying field $\mathbb{K}$, which in this case will be the field of real numbers $\mathbb{R}$. This element is what is known as the \textit{inner product}. To this effect, the metric tensor can be represented as a function, not necessarily linear, over each of the coordinates of the vectors it acts on. In geometric terms, the metric characterizes the underlying geometry of a vector space, by describing the projection of the underlying manifold of a non-Euclidean geometry to a Euclidean geometry $\mathbb{R}^n$ \cite{wald2010general}. The inner product between two vectors is informed by the metric in a precise way, and is representative of how the distance between two vectors should be calculated. 

A standard example consists of two unit vectors on a sphere, which is an $\mathbb{S}^2$ manifold that can be mapped onto $\mathbb{R}^3$. If the vectors are represented in spherical coordinates, which are a map from $\mathbb{S}^2$ to $\mathbb{R}^3$, the standard method of computing the angle between the vectors using Eq. (\ref{cosine}) will fail to give the correct value. The vectors need to be transformed by the appropriate non-linear metric to the Euclidean basis in $\mathbb{R}^3$ before a contraction of the coordinates can take place.
To illustrate this, take as an example a triangle drawn on the surface of a sphere $\mathbb{S}^2$. If it is projected onto a planisphere $\mathbb{R}^3$, a naive measurement of its internal angles will exceed the known 180 degrees, which corresponds to a change in the inner product between the vectors tangents to the triangle corners (see \citet{sphere} for a demonstration). To preserve this inner product, and thus recover the equivalence between a triangle on a spherical surface and a triangle on a Euclidean plane, the coordinates need to be properly transformed by the appropriate metric before they are contracted.

By the same token, we explore here the possibility that the shortcomings of the values obtained using cosine similarity when compared with human similarity ratings are not due to poor vector representations, but to a measure that fails to assess the distance between the vectors adequately. To test this hypothesis, we generalize the inner product of Eq. (\ref{euclideaninner}) to accommodate a larger class of relationships between vectors, modifying it using a metric represented by the distance matrix $d$, once a basis is assumed, that defines the inner product between two vectors as
\begin{equation} \label{generalinner}
\vec{n}_a \cdot_d \vec{n}_b=\sum_{ij} \vec{n}^i_a d^{ij} \vec{n}^j_b,
\end{equation}
where $\vec{n}^i_a$ is the $i$th component of $\vec{n}_a$. Using a metric of this form, the best we can achieve is a linear rescaling of the components of the vectors, which entails the existence of a non-orthogonal basis. The metric $d$ is required to be bilinear and symmetric, which is satisfied if
\begin{equation}
d^{sym}=B^TB,
\end{equation} such that Eq. (\ref{generalinner}) can be rewritten as

\begin{equation}\label{simmod}
\vec{n}_a \cdot_d \vec{n}_b = \left( B \vec{n}_a \right)^T \cdot  \left( B \vec{n}_b \right).
\end{equation}

We can thus learn the components of a metric for a certain set of vectors by fitting it to the goal of preserving a specified inner product. In the case of word similarity, the matrix $B$ can be learned supervised on human similarity judgments, towards the goal that a contextualized cosine similarity applied to a set of word embeddings, using Eq. (\ref{simmod}), returns the correct human assessment. An advantage of this approach is that the cosine is symmetric with respect to its inputs, which is a nice property that this extension preserves by requiring that symmetry of the metric.

\section{Methods} \label{methods}
%     \item contextualized dataset - different categories (...), different # word-pairs, scores 1-7

\begin{figure*}
    \centering
    \includegraphics[width=0.8\textwidth]{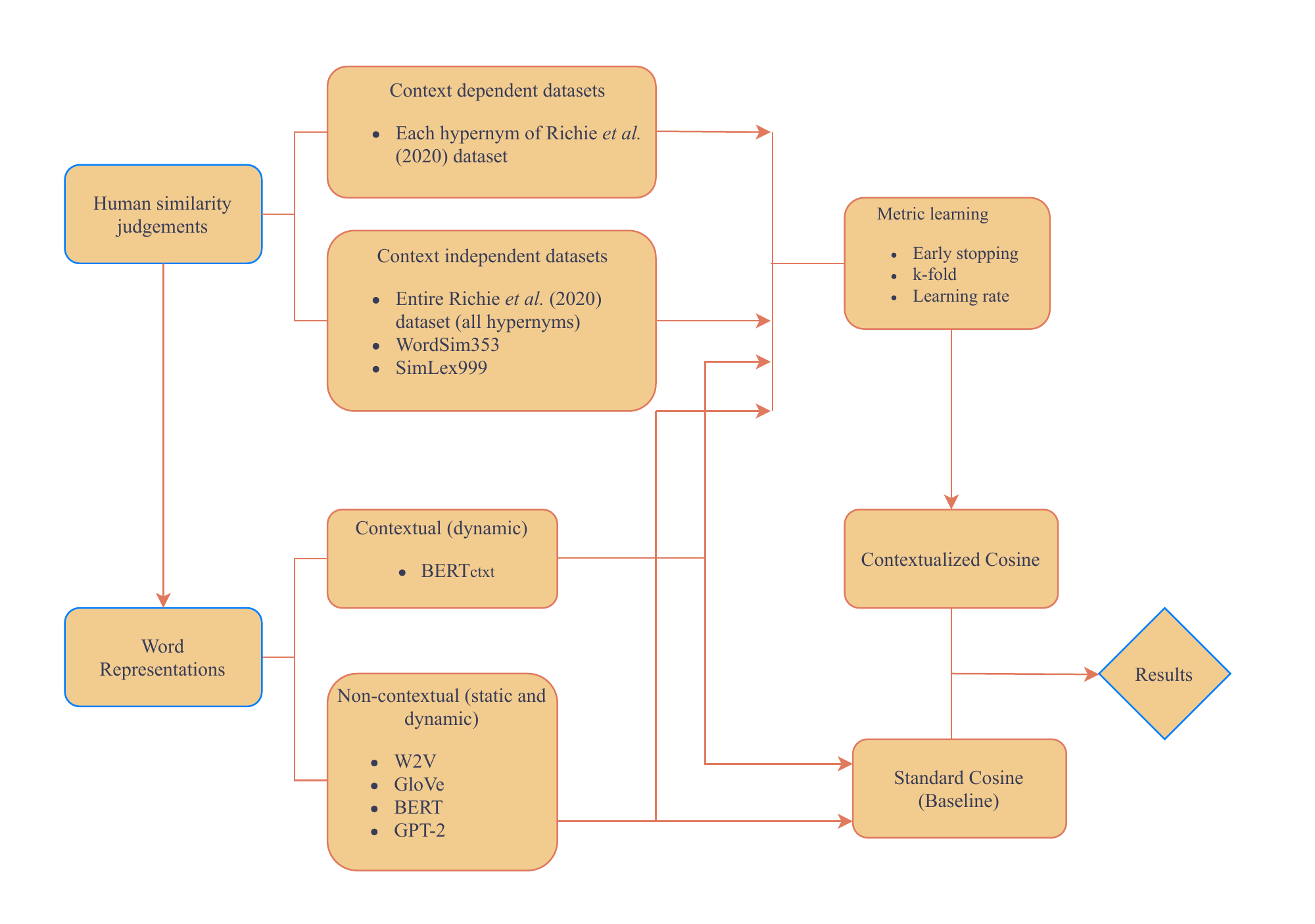}
    \caption{Schematic representation of the experiment leading up to the results in Tables \ref{tab:3} and \ref{tab:4}.}
    \label{flowchart}
\end{figure*}

The general outline of our experiment is as follows. First, we learn contextualized cosine similarity measures for related (contextualized) pairs of words, and afterwards for unrelated (non-contextualized) pairs of words. A schematic representation can be found in Fig. \ref{flowchart}. We then test whether these learned measures are transferable and provide improvements on word pairs that were not seen during training, when compared with the standard cosine similarity baseline.

\subsection{Datasets}
For a contextualized assessments of word similarity, we use the dataset of \citet{richie2020spatial}, where 365 participants were asked to judge the similarity between English word-pairs that are co-hyponyms of eight different hypernyms (Table \ref{tab:categories}). Participants were assigned a specific hypernym and were asked to rate the similarity between each co-hyponym pair from 1 to 7, with the highest rating indicating the words to be maximally similar. The number of annotators varies per hypernym, but each word-pair is rated by around 30 annotators, such that for the largest categories each annotator only saw a fraction of the totality of the word-pairs. As examples from the hypernym `Clothing', the word-pair `hat/overalls' was rated by 32 of the 61 annotators, resulting in an average similarity of 1.469, while `coat/gloves' had an average similarity rating of 3.281 and `coat/jacket' of 6.438, also by 32 annotators. The average similarity was computed for all word-pairs and rescaled to a value between 0 and 1, to be used as the target for supervised learning. 

Besides trying to fit a contextualized similarity measure to each hypernym, we also considered the entire all-hypernyms dataset, in order to test whether training on the hypernyms separately would result in a better cosine measure compared with when the hypernym information was disregarded.

To test whether similarity measures can be learned if the similarity of words is not assessed within a specific context, we use the WordSim-353 (WS353) \cite{finkelstein2001placing} and part of the SimLex-999 (SL999) \cite{hill2015simlex} datasets, where the word-pairs bear no specific semantic relation. From the SL999 dataset only the nouns were included, resulting in a dataset of 666 word-pairs. Additionally, we use these datasets to verify whether the similarity metric learned by training on the whole dataset of \citet{richie2020spatial} can be transferred to other, more general, datasets.

\begin{table}[t]\centering
    \caption{Number of words, word-pairs and human annotators per hypernym.}
    \small
    \label{tab:categories}
    \begin{tabular}{l|c|c|c}
         \hline
        Hypernym & Words & Pairs & Annotators \\
        \hline
        Birds       & 30 & 435 & 54 \\ 
        Clothing    & 29 & 406 & 61 \\
        Professions & 28 & 378 & 67 \\
        Sports      & 28 & 378 & 61 \\
        Vehicles    & 22 & 231 & 28 \\
        Fruit       & 21 & 210 & 31 \\
        Furniture   & 20 & 190 & 33 \\  
        Vegetables  & 20 &  190 & 30 \\
        \hline
        All       & 198 & 2418 & 365 \\
        \hline
    \end{tabular}
\end{table}

\subsection{Word embeddings}
To fine-tune the cosine similarity measure, we start from different pre-trained word representations. We do that for two classes of embeddings, static and dynamic.

Static embeddings were obtained from a pre-trained word2vec (W2V) model \cite{mikolov2013distributed} and a pre-trained GloVe model \cite{pennington2014glove}, each used to encode each word in the pair. Dynamic embeddings were obtained from two Transformers-based models, pre-trained BERT \cite{devlin2019bert} and GPT-2 models  \cite{radford2019language} (see Table \ref{repstable}). Here the representation of each word was taken to be the average representation of sub-word tokens when necessary, excluding the [CLS] and [SEP] tokens.

\begin{table}[t]
    \centering
    \label{tab:Representations}
    \resizebox{\columnwidth}{!}{%
    \begin{tabular}{l|c|c|c}
    \hline
    \textbf{Representation} & \textbf{Corpus} & \textbf{Corpus size} & \textbf{Dim} \\
    \hline
    word2vec & Google News & 100B & 300  \\
    \hline
    GloVe & GigaWord Corpus \& Wikipedia & 6B & 200  \\
    \hline
    BERT\textsubscript{base-uncased} & BooksCorpus \& English Wikipedia & 3.3B & 768  \\
    \hline GPT-2\textsubscript{medium} & 8 million web pages & $\sim$ 40 GB & 768  \\
    \hline
    \end{tabular}%
    }
    \caption{Pre-trained embeddings obtained from different source language models, with BERT and GPT-2 implemented using the Huggingface Transformers library.}\label{repstable}
\end{table}

\begin{table}[t]\centering
    \small
    \begin{tabular}{l|l}
         \hline
        Hypernym & Context words \\
        \hline
        Birds       & \texttt{small, migratory, other, } \\
        & \texttt{water, breeding} \\
        \hline
        Clothing    & \texttt{cotton, heavy, outer, winter,}\\
        & \texttt{leather} \\
        \hline
        Professions & \texttt{health, legal, engineering, } \\
        & \texttt{other, professional}  \\
        \hline
        Sports      & \texttt{youth, women, men, ea, boys}  \\
        \hline
        Vehicles    & \texttt{military, agricultural, motor,}\\ 
         & \texttt{recreational,  commercial}  \\
         \hline
        Fruit       & \texttt{citrus, summer, wild, sweet,} \\
        & \texttt{passion}  \\
        \hline
        Furniture   & \texttt{wood, furniture, modern,} \\
        & \texttt{antique, office} \\  
        \hline
        Vegetables  & \texttt{some, wild, root, fresh, green}  \\
        \hline
    \end{tabular}
    \caption{Five most likely words for masked token preceding hypernym token using BERT.}\label{tab:contexts}
\end{table}

%[max pulling]

\begin{figure}
\centering
\begin{subfigure}{0.4\textwidth}
    \includegraphics[width=\textwidth]{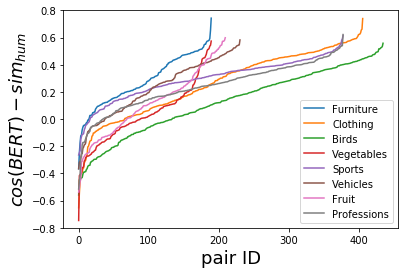}
    \caption{}
    \label{fig:first}
\end{subfigure}
\hfill
\begin{subfigure}{0.4\textwidth}
    \includegraphics[width=\textwidth]{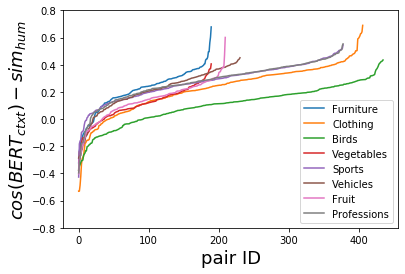}
    \caption{}
    \label{fig:second}
\end{subfigure}
\hfill
\begin{subfigure}{0.4\textwidth}
    \includegraphics[width=\textwidth]{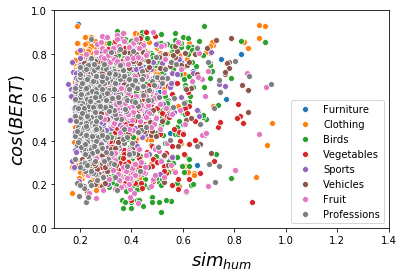}
    \caption{}
    \label{fig:third}
\end{subfigure}
\hfill
\begin{subfigure}{0.4\textwidth}
    \includegraphics[width=\textwidth]{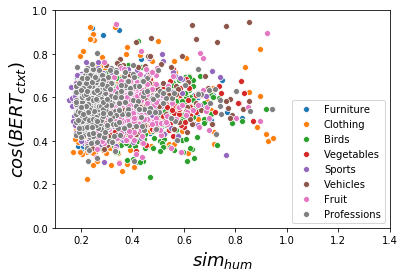}
    \caption{}
    \label{fig:fourth}
\end{subfigure}
        
\caption{Distributions of pairwise human similarity judgments $sim_{hum}$ and cosine similarity measures using either BERT representations ($\cos(\text{BERT})$) or contextualized BERT representations ($\cos(\text{BERT}_{ctxt})$). In (a) and (b) the absolute difference of scores, ordered per hypernym, is shown, while (c) and (d) represent the distribution of different similarity scores with respect to each other. Comparing the first two plots we can see a regularization effect by contextualizing the representations, and between the last two plots we can see a clustering effect.}
\label{fig:distribution}
\end{figure}

The token representations provided by the BERT model, as a bidirectional dynamic language model, can change depending on the surrounding context tokens. As such, additional contextualized embeddings were retrieved, BERT$_{ctxt}$, to test whether performance could be improved relative to the baseline cosine metric by using the hypernym information, as well as when compared with the hypernym cosine metric learned on non-contextualized representations. In this way we test whether leveraging the contextual information intrinsic to this dataset can in itself improve similarity at the baseline level, without the need of further training.

The contextualized vectors of BERT$_{ctxt}$ were obtained by first having BERT predict the five most likely adjectives that precede each hypernym using (\texttt{[MASK] <hypernym>}), and then using those adjectives to obtain five contextualized embeddings for each co-hyponym, subsequently averaged over.  Most of the predicted words were adjectives, and the few cases that were not were filtered out. For instance, for the category `Clothing', the most likely masked tokens were `cotton', `heavy', `outer', `winter' and `leather'. The contextualized representation of each hyponyms of `Clothing' was thus calculated as its average representation in the context of each of the adjectives, so that, for instance, for 'coat' we first obtained its contextualized representation in `cotton coat', `heavy coat', `outer coat', `winter coat', and `leather coat', performing a final averaging. The full list of context words can be found in Table \ref{tab:contexts}. Figs. \ref{fig:first} and \ref{fig:second} show that this transformation reduces the absolute extreme values of the difference between the values of the standard cosine similarity and the corresponding human similarity assessments, while regularizing the bulk of the differences closer to the desired value of 0. We tested other forms of contextualizing, such as (\texttt{<hypernym> is/are [MASK]}), but the resulting representations did not show as much improvement. 

The WS353 and SL999 datasets were only trained with non-contextualized embeddings, since we cannot obtain contextualized embeddings for the nouns in these datasets using the same method. For consistency, the models that were learned with contextualized representations were not tested on these datasets at the final step of our experiment.

%    learning part: model, parameters, possibly 
%     per category, all categories (1)
\subsection{Model}
A linear model was implemented on the PyTorch machine learning framework to learn the parameters of $B$, without a bias, such that a word initially represented by $\text{input}_a$ is transformed to $\text{input'}_a=B \text{input}_a$. The forward function of this model takes two inputs and returns
\begin{equation}\label{forward}
\frac{\left(\text{input'}_a \right)^T \cdot \text{input'}_b}{\sqrt{\left(\text{input'}_a\right)^T \cdot \text{input'}_1}\sqrt{\left(\text{input'}_b\right)^T \cdot \text{input}_b}},
\end{equation}  where $a$ and $b$ correspond to the indices of the words of a given word-pair\footnote{\url{ https://github.com/maradf/Contextualized-Cosine}}.

\subsection{Cross-validation} The number of co-hyponyms per hypernym is small when compared with the number of parameters in $B$ to be trained, which depends on the square of the dimension \textbf{Dim} of each representation. To ensure that the models did not overfit, a k-fold cross-validation was used during training \cite{raschka2015python}, which divided each dataset in k training sets and non-overlapping development sets. Additionally, early stopping of training was implemented in the event that the validation loss increased for ten consecutive epochs after it dropped below 0.1 \cite{bishop2006pattern}.

\begin{figure}[t]
    \centering
    \includegraphics[width=0.45\textwidth]{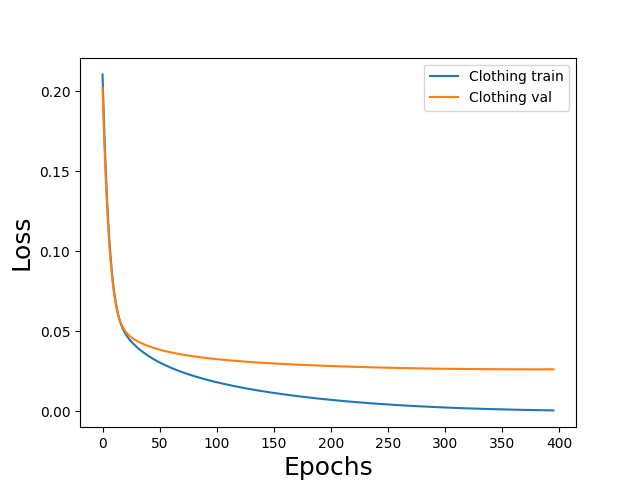}
    \caption{Example of learning curve, showing losses over epochs, from a fold training on the hypernym \textbf{Clothes} on the GloVe embeddings. In this case, training was stopped early at 397 epochs.}
    \label{fig:learningcurve}
\end{figure}

\subsection{Hyperparameter selection} Per each dataset $h$ (each hypernym, all hypernyms, WS353 or SL999) and learning rate $l_r$, k models $B^h_{i,l_r}$ were trained, with $i \in \{1,...,k\}$ and with k corresponding validation sets $val_i$. The training was done using two 16 cores (64 threads) Intel Xeon CPU at 2.1 GHz.

A fixed seed was used to find the best combination of the learning rate $l_r$ ($1\times 10^{-5}$, $1 \times 10^{-6}$, and $1\times 10^{-7}$) and the number of folds (5, 6 and 7) for the k-fold cross-validation. The regression to the best metric was done using the mean square error loss function and the Adam optimizer. The maximum number of training epochs was set to $500$, as most models converged at that point as per preliminary learning curve inspection (Fig.\ref{fig:learningcurve}). The implementation of early stopping resulted in \textit{de facto} variation of the number of epochs required to train each model.

\begin{table*}[h]
\begin{subtable}[h]{\textwidth}
\centering
\caption{Pearson correlations.}
\small
\begin{tabular}{lrrrrrrrrrrr}\toprule
\multirow{2}{*}{\textbf{Dataset (h)}} &\multicolumn{2}{c}{\textbf{BERT}} &\multicolumn{2}{c}{\textbf{BERT$_{ctxt}$}} &\multicolumn{2}{c}{\textbf{GPT-2}}  &\multicolumn{2}{c}{\textbf{word2vec}} &\multicolumn{2}{c}{\textbf{GloVe}} \\\cmidrule{2-11}
&Model &Base &Model &Base  &Model &Base &Model &Base &Model &Base \\\cmidrule{1-11}
Birds & \underline{\textbf{0.311}} & 0.098 & \underline{\textbf{0.316}} & 0.042 & \textbf{0.200} & -0.023 & \underline{\textbf{0.293}} & 0.213 & \textbf{0.215} & 0.194 \\
Clothing & \underline{\textbf{0.550}} & 0.141 & \underline{\textbf{0.515}} & 0.065 & \underline{\textbf{0.501}} & \underline{0.349} & \underline{\textbf{0.529}} & \underline{0.417} & \underline{\textbf{0.574}} & \underline{0.364} \\
Professions & \underline{\textbf{0.501}} & 0.193 & \underline{\textbf{0.601}} & 0.073 & \underline{\textbf{0.651}} & \underline{0.542} &  \underline{\textbf{0.635}} & \underline{0.566} & \underline{0.529} & \underline{0.529} \\
Sports & \underline{\textbf{0.452}} & 0.175 & \underline{\textbf{0.543}} & 0.139 & \underline{\textbf{0.556}} & \underline{0.324} &  \underline{\textbf{0.532}} & \underline{0.418} & \underline{\textbf{0.580}} & \underline{0.386} \\
Vehicles & \underline{\textbf{0.496}} & 0.218 & \underline{\textbf{0.616}} & 0.123 & \underline{\textbf{0.645}} & \underline{0.385} &  \underline{\textbf{0.738}} & \underline{0.719} & \underline{\textbf{0.703}} & \underline{0.567} \\
Fruit & \textbf{0.315} & 0.016 & \textbf{0.378} & -0.037 & \textbf{0.333} & 0.203 &  \textbf{0.361} & 0.239 & \underline{\textbf{0.571}} & 0.392 \\
Furniture & \textbf{0.353} & -0.018 & \underline{\textbf{0.539}} & -0.035 & \underline{\textbf{0.568}} & 0.399 &  \underline{\textbf{0.368}} & 0.333 & \underline{\textbf{0.470}} & \underline{0.462} \\
Vegetables & \textbf{0.211} & -0.059 & \textbf{0.293} & -0.044 & \textbf{0.378} & 0.144 &  \underline{\textbf{0.577}} & 0.281 & \underline{\textbf{0.562}} & 0.290 \\

All hypernyms & \underline{\textbf{0.434}} & 0.100 & \underline{\textbf{0.542}} & 0.040 & \underline{\textbf{0.508}} & 0.287 & \underline{\textbf{0.483}} & 0.400 & \underline{\textbf{0.539}} & 0.397 \\
WordSim-353 & \underline{\textbf{0.517}} & 0.238 & - & - & \underline{\textbf{0.651}} & \underline{0.647} &  \underline{0.637} & \underline{0.654} & \underline{\textbf{0.622}} & \underline{0.568} \\
SimLex-999 & \underline{\textbf{0.403}} & 0.161 & - & - & \underline{\textbf{0.555}} & \underline{0.504} &  \underline{\textbf{0.495}} & \underline{0.455} & \underline{\textbf{0.510}} & \underline{0.408} \\
\bottomrule
\end{tabular}
\end{subtable}
\begin{subtable}[h]{\textwidth}
\centering
\caption{Spearman correlations.}
\small
\begin{tabular}{lrrrrrrrrrrr}\toprule
\multirow{2}{*}{\textbf{Dataset (h)}} &\multicolumn{2}{c}{\textbf{BERT}} &\multicolumn{2}{c}{\textbf{BERT$_{ctxt}$}} &\multicolumn{2}{c}{\textbf{GPT-2}}  &\multicolumn{2}{c}{\textbf{word2vec}} &\multicolumn{2}{c}{\textbf{GloVe}} \\\cmidrule{2-11}
&Model &Base &Model &Base  &Model &Base &Model &Base &Model &Base \\ \cmidrule{1-11}
Birds & \underline{\textbf{0.260}} & 0.102 & \underline{\textbf{0.299}} & 0.052 & \textbf{0.190} & -0.054 &  \underline{\textbf{0.250}} & 0.211 & \textbf{0.238} & 0.201
 \\
Clothing & \underline{\textbf{0.436}} & 0.184 & \underline{\textbf{0.467}} & 0.059 & \underline{\textbf{0.445}} & 0.276 & \underline{\textbf{0.510}} & \underline{0.414} & \underline{\textbf{0.513}} & \underline{0.384} \\
 Professions & \underline{\textbf{0.501}} & 0.248 & \underline{\textbf{0.578}} & 0.170 & \underline{\textbf{0.560}} & \underline{0.473} &  \underline{\textbf{0.518}} & \underline{0.410} & \underline{0.482} & \underline{0.486} \\
 Sports & \underline{\textbf{0.391}} & 0.174 & \underline{\textbf{0.526}} & 0.142 & \underline{\textbf{0.540}} & 0.291 &  \underline{\textbf{0.458}} & \underline{0.339} & \underline{\textbf{0.478}} & \underline{0.325} \\
 Vehicles & \underline{\textbf{0.518}} & 0.238 & \underline{\textbf{0.601}} & 0.056 & \underline{\textbf{0.626}} & 0.288 &  \underline{\textbf{0.709}} & \underline{0.687} & \underline{\textbf{0.680}} & \underline{0.596} \\
Fruit & \textbf{0.265} & -0.014 & \underline{\textbf{0.333}} & -0.103 & \textbf{0.365} & 0.173 & \textbf{0.368} & 0.277 & \underline{\textbf{0.491}} & 0.342 \\
Furniture & \textbf{0.353} & -0.032 & \underline{\textbf{0.491}} & -0.120 & \underline{\textbf{0.527}} & 0.393 &  \underline{\textbf{0.442}} & \underline{0.402} & \textbf{0.464} & \underline{0.451} \\
Vegetables & \textbf{0.217} & -0.028 & \textbf{0.305} & 0.015 & \underline{\textbf{0.363}} & 0.089 &  \underline{\textbf{0.587}} & 0.290 & \underline{\textbf{0.528}} & 0.228 \\
All hypernyms & \underline{\textbf{0.407}} & 0.111 & \underline{\textbf{0.504}} & 0.034 & \underline{\textbf{0.504}} & 0.242 &  \underline{\textbf{0.446}} & 0.379 & \underline{\textbf{0.477}} & 0.377 \\
WordSim-353 & \underline{\textbf{0.543}} & 0.267 & - & - & \underline{\textbf{0.715}} & \underline{0.705} &  \underline{0.675} & \underline{0.701} & \underline{\textbf{0.624}} & \underline{0.579} \\
SimLex-999 & \underline{\textbf{0.416}} & 0.180 & - & - & \underline{\textbf{0.566}} & \underline{0.513} &  \underline{\textbf{0.475}} & \underline{0.445} & \underline{\textbf{0.500}} & \underline{0.374} \\
\bottomrule
\end{tabular}
\end{subtable}
\caption{Best correlation scores between human similarity judgments and similarity scores found by the trained model, compared with baseline cosine metric values of the same hyperparameters. The underlined correlation values are the statistical significant values with a p $<$ 0.05, and the bold values correspond to model correlations that were higher than base correlations.}\label{tab:3}
\end{table*}

\begin{table*}[h]
\begin{subtable}[h]{\textwidth}
\centering
\caption{Pearson correlations.}
\small
\begin{tabular}{lrrrrrrrrrrr}\toprule
\multirow{2}{*}{\textbf{Dataset (h)}} &\multicolumn{2}{c}{\textbf{BERT}} &\multicolumn{2}{c}{\textbf{BERT$_{ctxt}$}} &\multicolumn{2}{c}{\textbf{GPT-2}}  &\multicolumn{2}{c}{\textbf{W2V}} &\multicolumn{2}{c}{\textbf{GloVe}} \\\cmidrule{2-11}
&$\%$ &$l_r$, k &$\%$ &$lr$, k  &$\%$ &$lr$, k &$\%$ &$lr$, k &$\%$ &$lr$, k \\\cmidrule{1-11}
Birds &	217	& $10^{-6}$,5 & 652 & $10^{-6}$,	5
&	\textbf{770} & $10^{-5}$,	5
 &	38 & $10^{-5}$,	5
 &	11 & $10^{-5}$,	7
 \\
Clothing &	290	& $10^{-6}$,5 & \textbf{692} & $10^{-6}$,	6
 &	44 & $10^{-5}$,	6
 &	27 & $10^{-5}$,	7
 &	58 & $10^{-6}$,	5
\\
Professions &	160 & $10^{-6}$,	5 &	\textbf{723} & $10^{-6}$,	6
 &	20 & $10^{-5}$,	5
 &	12 & $10^{-5}$,	7
 &	0 & $10^{-5}$,	5
 \\
Sports & 158 & $10^{-5}$,	6
 & \textbf{291} & $10^{-6}$,	6
 &	72 & $10^{-5}$,	6
 &	27 & $10^{-5}$,	6
 &	50 & $10^{-6}$,	7
\\
Vehicles &	128 & $10^{-6}$,	6
 &	\textbf{401} & $10^{-5}$,	7
 &	68 & $10^{-5}$,	5
 &	3 & $10^{-5}$,	5
 &	24 & $10^{-6}$,	6
\\
Fruit &	\textbf{1869} & $10^{-5}$, 7
 &	922 & $10^{-6}$,	6
 &	64 &  $10^{-5}$,	7
 &	51 &  $10^{-6}$,	5
 &	46 & $10^{-7}$,	7
 \\ 
Furniture &	\textbf{1861}	 & $10^{-5}$,	7
 & 1440 & $10^{-6}$,	6
 & 42 & $10^{-5}$,	7
 & 11 & $10^{-5}$,	6
 & 	2 & $10^{-5}$,	6
\\
Vegetables &	258 & $10^{-5}$,	7
 &	\textbf{566} &  $10^{-6}$,	6
 &	163 & $10^{-5}$,	5
 &	105 & $10^{-6}$,	7
 &	94 & $10^{-6}$,	5
 \\
All &	334 & $10^{-5}$,	5
 &	\textbf{1255} & $10^{-6}$,	7
 & 77 & $10^{-5}$,	6
 &	21 & $10^{-5}$,	6
 &	36 & $10^{-7}$,	6
\\
WordSim-353 & 	\textbf{117} & $10^{-6}$,	7
 &	- & - & 	1 & $10^{-5}$,	7
 &	-3 & $10^{-5}$,	6
 &	10 & $10^{-5}$,	5
 \\
SimLex-999 &	\textbf{150} & $10^{-6}$,	7
 &	- & - &	10 & $10^{-5}$,	6
 &	9	& $10^{-5}$,	6
 & 25 & $10^{-6}$,	5
\\
\bottomrule
\end{tabular}
\end{subtable}
\begin{subtable}[h]{\textwidth}
\centering
\caption{Spearman correlations.}
\small
\begin{tabular}{lrrrrrrrrrrr}\toprule
\multirow{2}{*}{\textbf{Dataset (h)}} &\multicolumn{2}{c}{\textbf{BERT}} &\multicolumn{2}{c}{\textbf{BERT$_{ctxt}$}} &\multicolumn{2}{c}{\textbf{GPT-2}}  &\multicolumn{2}{c}{\textbf{W2V}} &\multicolumn{2}{c}{\textbf{GloVe}} \\\cmidrule{2-11}
&$\%$ &$lr$, k &$\%$ &$lr$, k  &$\%$ &$lr$, k &$\%$ &$lr$, k &$\%$ &$lr$, k \\ \cmidrule{1-11}
Birds &	155 & $10^{-6}$,	5
 &	\textbf{475} & $10^{-6}$,	5
 &	252 & $10^{-5}$,	7
 &	18 & $10^{-5}$,	5
 &	18 & $10^{-7}$,	5
\\
Clothing &	137 & $10^{-6}$,	5
 &	\textbf{692} & $10^{-6}$,	6
 &	61	& $10^{-5}$,	7
 & 23 & $10^{-5}$,	7
 &	34 & $10^{-6}$,	5
\\
Professions &	102 & $10^{-6}$,	7
 &	\textbf{240} & $10^{-6}$,	5
 &	18 & $10^{-5}$,	5
 &	26 & $10^{-5}$,	7
 &	-1 & $10^{-7}$,	6
\\
Sports & 125 & $10^{-5}$,	6
 & \textbf{270} & $10^{-6}$,	6
 & 86 & $10^{-5}$,	6
 &	35 & $10^{-5}$,	6
 & 47 & $10^{-6}$,	6
\\
Vehicles &	118 & $10^{-6}$,	6
 &	\textbf{973} & $10^{-6}$,	6
 &	117 & $10^{-5}$,	7
 &	3 & $10^{-5}$,	5
 & 14 & $10^{-6}$,	6
\\
Fruit &	\textbf{1793} & $10^{-6}$,	7
 & 223 & $10^{-6}$,	6
 & 111 & $10^{-5}$,	6
 & 33 & $10^{-6}$,	6
 &	44 & $10^{-7}$,	7
\\
Furniture &	\textbf{1003} & $10^{-6}$,	6
 & 309 & $10^{-6}$,	5
 & 34 & $10^{-5}$,	5
 &	10 & $10^{-5}$,	6
 & 3 & $10^{-6}$,	7
\\
Vegetables & 675 & $10^{-5}$,	7
 & \textbf{1933} & $10^{-6}$,	6
 & 308 & $10^{-5}$,	5
 & 102 & $10^{-6}$,	7
 & 132 & $10^{-6}$,	5
\\
All hypernyms &	267 & $10^{-5}$,	5
 &	\textbf{1382} & $10^{-6}$,	7
 & 108 & $10^{-5}$,	6
 & 18 & $10^{-5}$,	6
 & 27 & $10^{-6}$,	5
 \\
WordSim-353 & \textbf{103}  & $10^{-6}$,	5
 & - & - & 1 & $10^{-5}$,	7
 &	-4 & $10^{-6}$,	5
 &	8 & $10^{-5}$,	5
\\
SimLex-999 & \textbf{131} & $10^{-6}$,	7
 & - & - &	10 & $10^{-5}$,	6
 &	7 & $10^{-5}$,	6
 & 34 & $10^{-6}$,	5
\\
\bottomrule
\end{tabular}
\end{subtable}
\caption{Change ($\%$) in correlation from Table \ref{tab:3}, given by $(|\text{Model}|-|\text{Base}|)/|\text{Base}|$, at corresponding best hyperparameters ($lr$, k). Values in bold indicate the highest increase on a given dataset.}\label{tab:4}
\end{table*}

\subsection{Testing the model}

Each one of the $B^h_{i,l_r}$ models was tested on the corresponding holdout validation set $val_i$, resulting in two correlation scores between the models' predicted similarity scores and the human judgment scores: a Pearson correlation score $r^{h}_{i,l_r}(val^h_i)$ and a Spearman correlation score $\rho^{h}_{i,l_r}(val^h_i)$. A final score per k and $l_r$ was calculated using the average performance on the validation sets as

\begin{align}\label{modelcorrs1}
 &   r^{h}_{k,l_r}= \frac{1}{k}\sum_{i=1}^k r^{h}_{i,l_r}(val^h_i), \\
 &   \rho^{h}_{k,l_r}= \frac{1}{k}\sum_{i=1}^k \rho^{h}_{i,l_r}(val^h_i). \label{modelcorrs2}
\end{align}

The baseline results were obtained in a similar form, but with the model $B^{std}$ corresponding to the identity matrix, returning the standard cosine similarity rating as

\begin{align}
 &   r^{h,std}_{k}= \frac{1}{k}\sum_{i=1}^k r^{std}(val^h_i), \label{baseline11} \\
 &   \rho^{h,std}_{k}= \frac{1}{k}\sum_{i=1}^k \rho^{std}(val^h_i). \label{baseline12}
\end{align}

%     
%    non contextualized datasets (wordsim, simlex99)
%        non context reps, learn, comparing baseline, 
%        use these datasets as test set for all categories model learned from (1)
%\end{itemize}

The model results shown in Table \ref{tab:3} correspond to the best correlation values obtained using Eqs. (\ref{modelcorrs1}) and (\ref{modelcorrs2}), with the baselines given as in Eqs. (\ref{baseline11}) and (\ref{baseline12}). The hyperparameters corresponding to the best results can be found in Table \ref{tab:4}, along with the relative change in correlation performance. As the seed was fixed, the differences in performance achieved by models trained on each hypernym and on all-hypernyms of the contextualized dataset were not due to randomization errors. The final correlation per fold on the entire all-hypernyms dataset was found by first calculating the correlation per hypernym and then averaging over all eight hypernyms.  

To test the transferability of the metric learned on the all-hypernyms dataset to other datasets, the model that returned the best correlation scores on the validation datasets of the all-hypernyms dataset was tested on the entire WS353 and SL999 datasets. As the best performing model consists in fact of k models, each one of these was tested on the entire datasets, as

\begin{align} 
 &   r^{h,test}_{k,l_r}= \frac{1}{k}\sum_{i=1}^k r^{All-hyp}_{i,l_r}(test^h), \\
 &   \rho^{h,test}_{k,l_r}= \frac{1}{k}\sum_{i=1}^k \rho^{All-hyp}_{i,l_r}(test^h),
\end{align} with $h \in \{ \text{WS353}, \text{SL999}\}$.
% \end{align}

The baselines for these results were obtained by applying $B^{std}$ to the entire WS353 and SL999 datasets as

\begin{align}
 &   r^{h,std}= r^{std}(test^h), \label{baseline21} \\
 &   \rho^{h,std}= \rho^{std}(test^h) \label{baseline22}.
\end{align} As the correlation functions are not linear, the results from Eqs. (\ref{baseline11}) and (\ref{baseline12}) for the WS353 and SL999 datasets are expected to differ from those obtained using Eqs. (\ref{baseline21}) and (\ref{baseline22}) for the same datasets.

\section{Results} \label{results}
The validation results on Table \ref{tab:3} show consistent improvements over the baselines, with statistical significance. This confirms that the modification introduced to the cosine measure worked in a principled way, and consistent with the results found by \citet{richie2021similarity}. On the individual hypernym datasets, `Vehicles' showed the best correlations, except for the Pearson correlation in GPT-2, in spite of not being the largest hypernym dataset. On the contrary, the smallest categories showed the lowest correlations. In general, the relative performance of hypernyms according to the baselines extends to the model correlations, although with better performance. With some exceptions, mainly in the `Birds' hypernym, the best performing representation was GPT-2, followed by W2V, but the relative increase as shown in Table \ref{tab:4} was clearly superior for the dynamic representations. An important observation that we make is that the model trained on all hypernyms had a better performance than the average performance on the individual hypernyms. As the seed was fixed, this means that the performance on the hypernym-specific validation sets increased if at training time the models saw more examples, from different categories, indicating that a similarity relationship was learned and transferred across different contexts. Improvements over baseline also took place if a metric was learned on datasets where the word pairs did not share a context, as was the case with WS353 and SL999, but the percentual increase was lower, as seen in Table \ref{tab:4}.

Comparing the results of BERT contextualized and non-contextualized, the baseline values of the contextualized representations were worse than those obtained with the contextualized embeddings, although without statistical significance, while the improvement after training was consistently better and significant for all datasets with the contextualized representations. Figs. \ref{fig:third} and \ref{fig:fourth}, show that the distribution of points using the contextualized embeddings is more concentrated and collinear, making it more likely that a metric that acts in the same way for all points in the dataset will rotate and rescale them into a positive correlation. The percentual increases also show that BERT contextualized had the greatest increases from before to after training, suggesting that there was a cumulative effect in considering the context both in the representations and in the similarity measure.

Table \ref{tab:5} shows the results of applying the best model learned on all hypernyms to the WS353 and SL999 datasets. The baseline values for the static representations are comparable with the existing literature \cite{toshevska2020comparative}. We see that our model was capable of improving on the correlation scores on the datasets, for some representations. Although the improvements did not happen across the board, they show clear evidence that the notion of similarity in the form of a modified cosine measure can be learned in one dataset and applied with positive results to an independent dataset.
\\

\begin{threeparttable}[t]
\centering
\small
\par
\begin{tabular}{lrrrrrr}\toprule
\multicolumn{2}{c}{\multirow{2}{*}{}} &\multicolumn{2}{c}{\textbf{Pearson}} &\multicolumn{2}{c}{\textbf{Spearman}} \\\cmidrule{3-6}
& &WS353 &SL999 &WS353 &SL999 \\\cmidrule{1-6}
\multirow{2}{*}{\textbf{BERT}} & Model & \textbf{0.487} & \textbf{0.375} & \textbf{0.519} & \textbf{0.384} \\
&Base & 0.239 &  0.151 & 0.267 & 0.172 \\
\multirow{2}{*}{\textbf{GPT-2}} & Model & 0.635 & \textbf{0.507} & 0.676 & 0.513 \\
&Base & 0.647 & 0.504 & 0.709 & 0.520 \\
\multirow{2}{*}{\textbf{W2V}} & Model & 0.613 & 0.472 & 0.632 & \textbf{0.457} \\
&Base & 0.653 & 0.460 & 0.700 & 0.452 \\
\multirow{2}{*}{\textbf{GloVe}} & Model & \textbf{0.593} & \textbf{0.431} & 0.558 & \textbf{0.392} \\
&Base & 0.578 & 0.408 & 0.578 & 0.376 \\
\bottomrule \\

\textbf{SOTA} &  &  0.704	 &  	0.658 &  0.828  &  0.76 
\end{tabular}
\caption{Best model trained on all hypernyms, tested on SimLex-999 and WordSim-353 datasets.  Bold values indicate correlation scores above baseline, and underlining indicates statistical significance. State of the art from \citet{recski2016measuring, dobo2020comprehensive, speer2017conceptnet, banjade2015lemon}.}\label{tab:5}
\end{threeparttable}

\section{Conclusion and Outlook} \label{conclusion}

In this paper we tested whether a contextualized notion of cosine similarity could be learned, improving the similarity not only of the results for the datasets where it was learned, but of unrelated similarities. We showed that this metric improved the correlations above baseline, and that, when learned on a contextualized similarity dataset, it had an advantage when compared to one learned on a dataset with unrelated word-pairs. We furthermore showed that this framework has the potential to generalize the notion of similarity to word-pairs it has not seen during training. An important future research line towards interpretability consists in understanding the properties of the metrics that yielded the best results, particularly in identifying the distinctive features of the best metrics, such as their eigensystems. Other further directions include applying these metrics to distributional compositional contractions, including with dependency enhancements \cite{kogkalidis2019constructive}, testing this framework on larger contextualized datasets and trying out more complex, non-linear, metric forms.

\section*{Acknowledgements}
All authors would like to thank Juul A. Schoevers for contributions made during the early stages of the project. A.D.C. would like to thank Gijs Wijnholds, Konstantinos Kogkalidis, Michael Moortgat and Henk T.C. Stoof for the many exchanges during this research. This work is supported by the UU Complex Systems Fund, with special thanks to Peter Koeze.

%Discussion on parameters
%The experiment failed to show that a metric trained on contextualized datasets improved the similarity results when compared to baselines. Possible methodological improvements include a more systematic evaluation of hyperparameters, tailored to each representation, since the final results seem to work better on the BERT representations as compared to GloVe, W2V and GPT-2. Contextualized GPT-2 representations were too large to be trained on all hypernyms and were therefore left out of the scope of this research. We conclude that the model was too simple considering the size of the dataset, but possibly better results would be obtained on larger co-hyponym datasets.

%\section*{Acknowledgements}

\bibliography{acl2020.bib}
\bibliographystyle{acl_natbib}

\appendix

\end{document}